\documentclass [12pt]{iopart}
\usepackage {iopams}
\usepackage{setstack}
\usepackage{amstext}
\usepackage{amsfonts}
\usepackage{graphicx}
\usepackage{subfigure}
\usepackage{array}
\usepackage{appendix}
\usepackage{mathrsfs}
\usepackage{color}
\usepackage{adjustbox,lipsum}
\usepackage{upgreek}
\usepackage{algorithm}
\usepackage{algpseudocode}
\usepackage{makecell}
\usepackage{adjustbox,lipsum}

\newcommand{\bra}[1]{\left<#1\right|}
\newcommand{\ket}[1]{\left|#1\right>}
\newcommand{\abs}[1]{\left|#1\right|}

\newtheorem{theorem}{Theorem}
\newtheorem{definition}{Definition}
\newtheorem{corollary}{Corollary}
\newtheorem{proposition}{Proposition}
\newtheorem{remark}{Remark}
\newtheorem{conjecture}{Conjecture}

\begin{document}
\title[]{Ensuring superior learning outcomes and data security for authorized learner}

\author{Jeongho~Bang$^1$, Wooyeong~Song$^{2}$, Kyujin~Shin$^{3,4}$, and Yong-Su~Kim$^{4,5}$}

\address{$^1$ Institute for Convergence Research and Education in Advanced Technology, Yonsei University, Seoul 03722, Republic of Korea}
\address{$^2$ Quantum Network Research Center, Korea Institute of Science and Technology Information (KISTI), Daejeon 34141, Republic of Korea}
\address{$^3$ Materials Research \& Engineering Center (MREC), Advanced Vehicle Platform Division, Hyundai Motor Company, Uiwang 16082, Republic of Korea}
\address{$^4$ Center for Quantum Technology, Korea Institute of Science and Technology (KIST), Seoul 02792, Republic of Korea}
\address{$^5$ Division of Quantum Information, KIST School, Korea University of Science and Technology, Seoul 02792, Republic of Korea}

\vspace{10pt}

\begin{indented}
\item {\em Correspondence and requests for materials should be addressed to J.B and Y.S.K}
\end{indented}

\ead{\mailto{jbang@yonsei.ac.kr} and \mailto{yong-su.kim@kist.re.kr}}

\begin{abstract}
The learner's ability to generate a hypothesis that closely approximates the target function is crucial in machine learning. Achieving this requires sufficient data; however, unauthorized access by an eavesdropping learner can lead to security risks. Thus, it is important to ensure the performance of the ``authorized'' learner by limiting the quality of the training data accessible to eavesdroppers. Unlike previous studies focusing on encryption or access controls, we provide a theorem to ensure superior learning outcomes exclusively for the authorized learner with quantum label encoding. In this context, we use the probably-approximately-correct (PAC) learning framework and introduce the concept of learning probability to quantitatively assess learner performance. Our theorem allows the condition that, given a training dataset, an authorized learner is guaranteed to achieve a certain quality of learning outcome, while eavesdroppers are not. Notably, this condition can be constructed based only on the authorized-learning-only measurable quantities of the training data, i.e., its size and noise degree. We validate our theoretical proofs and predictions through convolutional neural networks (CNNs) image classification learning.
\end{abstract}

\maketitle

\section{Introduction}\label{sec:1}

Connecting the two state-of-the-art science fields, machine learning and quantum computing, has been of keen interest very recently. Starting with understanding what quantum advantages are achievable in machine learning to how to implement, a variety of studies has currently been conducted---hence, a subfield, so-called the quantum machine learning (QML), has emerged~\cite{Biamonte2017,Ciliberto2018}. One of the celebrated results in QML is the achievement of an exponential computing speedup in data classification by using useful quantum linear algebra kernels~\cite{Rebentrost2014}. This result has been generalized other machine learning tasks requiring the linear optimization, such as linear regression~\cite{Schuld2016,Wang2017}, principal component analysis~\cite{Lloyd2014}, etc. Such a generalized method for QML includes the strong theoretical proofs of the quantum computational learning speedup\footnote{However, although promising, several controversies also exist in developing QML based on the quantum linear-algebra kernels because it appears to be impractical to realize the quantum speedup without accessing a {\em largely-}superposed sample, or equivalently, without using a imaginary quantum gadget, quantum random-access memory. For more details, refer to Refs.~\cite{Aaronson2015,Arunachalam2015,Tang2021}.}.

Meanwhile, the field of QML has shifted its focus to studying the properties of data described in Hilbet-space; in other words, (so-called) the quantum data has become important\footnote{Note that while by ``quantum data'' here we mean the quantum state into which the classical data is encoded, it is not a general-purpose terminology in QML.}. Such a development trend is unsurprising since the machine learning is essentially a data-driven process. One significant recent achievement is the discovery that some eligible data representations in the quantum Hilbert-space, such as quantum feature map, offer the quantum computational advantages~\cite{Schuld2019,Havlivcek2019}. Since then, subsequent researches have continued to explore the various QML advantages in terms of the quantum data-encoding~\cite{Lloyd2020,Weigold2021}: in short,
\begin{eqnarray}
\{(\mathbf{x}, c(\mathbf{x}))\} \to \ket{\Psi (\mathbf{x}, c(\mathbf{x}))},
\label{eq:q_data_en}
\end{eqnarray}
where $\mathbf{x} \in {\cal X}$ denotes a user-recognizable classical data and $c \in C$ is a target map.

From another perspective, there have been several studies exploring the properties of the security in QML. Here, the security in the learning refers to the safeguarding of a client's confidential data and learning outcomes from any external intruders. This scenario, i.e., in which more than one learning party is involved, is natural in machine learning and becomes a useful interface between the security, computation, and machine learning. In a few early studies and subsequent works, it has been shown that a client can learn a task securely by leveraging a server at a distant place~\cite{Bang2015,Sheng2017}. After that, the efforts are underway to contextualize the ideas devised for specific applications into general principles. For example, the security-related QML algorithms, such as anomaly detection, have been studied in a data-networked setting~\cite{Liu2018,Du2021,Llorens2024}. In Ref.~\cite{Song2021a}, a security condition is defined by the theoretical bounds of the learning sample complexity. Recently, a secure quantum pattern encoding has been studied in the framework of the continuous-variable system~\cite{Harney2022}. However, there remains a lack of studies that leverage the power of the quantum-encoded data in terms of the machine learning security.

Building upon previous research and in the line with Ref.~\cite{Song2021a}, we investigate the theoretical conditions under which an authorized learner exclusively achieves superior learning outcomes. To this end, we introduce the concept of learning probability within the computational learning framework known as probably-approximately-correct (PAC) learning~\cite{Valiant1984,Langley1996}. This model setting enables us to quantitatively measure the performance of authorized and eavesdropping learners. By utilizing a classical-quantum hybrid data encoding~\cite{Lee2019,Song2021b}, dubbed here as quantum label encoding, we present a theorem proving that a certain level of learning outcome quality is guaranteed exclusively for the authorized learner and not for the eavesdropping learners. Our study extends beyond theoretical proof to a practical application, the image classification using convolutional neural network (CNN). This demonstration confirms that a specific quality of learning outcomes, attainable only by the authorized learner, can indeed be observed in practical tasks. We expect our study to offer a framework for analyzing the learning performances in quantum secure learning scenarios, with a focus solely on the properties of the quantum training data.

\section{Authorized-learner attainable learning probability}\label{sec:2}

In a broad sense, the learning is defined as a process of returning a hypothesis function $h \in H$ close to a given target function, called concept, $c \in C$ which maps the inputs $\mathbf{x}$ to their corresponding labels $c(\mathbf{x})$. Here, an important assumption is that a learner, say $\mathscr{L}$, can access to a {\em finite} set $S$ of the training data. The difference of $h$ to $c$ is evaluated by using an error function $R(c,h)=\frac{1}{\abs{T}} \sum_{\mathbf{x} \in T} \mathbb{I}_{c(\mathbf{x}) \neq h(\mathbf{x})}$, where $T$ is the test dataset and $\mathbb{I}_{\omega}$ is the indicator of the event $\omega$~\cite{Langley1996}. 

In such a framework, the classification has frequently been studied due to its wide applicability. In particular, one of the beauties of the classification problem is that it allows the data-driven analysis~\cite{Kotsiantis2011}. Thus we consider the classification problem in this study. The classification is defined as below:
\begin{definition}
Given a fixed $H$, assume that $\mathscr{L}$ can access the training data such that 
\begin{eqnarray}
\Theta=\{ (\mathbf{x}, y=c(\mathbf{x})) | \mathbf{x} \in {\cal X}, c \in C \}.
\label{eq:l_data}
\end{eqnarray}
Then, the task is to find the best $h \in H$ that has a small $R(c,h)$. Here, $\mathbf{x}$ represents an input of arbitrary form and $y = c(\mathbf{x}) \in \{0, 1\}$.
\end{definition}

\subsection{Probably-Approximately-Correct learning} 

In computational learning theory, we need useful measures of the quality of the learner $\mathscr{L}$. Here, we introduce the model of probably-approximately-correct (PAC) learning, which provides a notion of sample complexity to evaluate the size of the training data (i.e., $\abs{\Theta}$ in our case) required for $\mathscr{L}$ to learn a family of $C$. Firstly, we define a PAC learner as~\cite{Valiant1984}
\begin{definition}
For any concept class $C$ and any dataset $\Theta$ satisfying $\abs{\Theta} \le \text{poly}(\frac{1}{\epsilon}, \frac{1}{\delta}, \abs{C})$, if an $\epsilon$-approximated hypothesis $h$ is returned with a probability of more than $1-\delta$, namely if the following is satisfied,
\begin{eqnarray}
Pr(R(h, c) \le \epsilon) \ge 1-\delta,
\label{eq:def_PAC}
\end{eqnarray}
the concept class $C$ is said to be ``PAC-learnable'' and $\mathscr{L}$ is called ``$(\epsilon, \delta)$-PAC learner.'' Here, $\epsilon$ and $1-\delta$ are known as the inaccuracy and confidence, respectively.
\end{definition}
In the PAC learning model, the following theorem has been proven:
\begin{theorem}
$\mathscr{L}$ can become a $(\epsilon, \delta)$-PAC learner iff 
\begin{eqnarray}
\abs{\Theta} \ge M_b = \frac{1}{\epsilon}\ln{\frac{{\abs{H}}}{\delta}},
\label{eq:M_b_no_eta}
\end{eqnarray}
where $\abs{H}$ is given as a complexity of the space $H$, often-called the model complexity. Here, we consider that $\mathscr{L}$ has a finite model parameters.
\label{thm:1}
\end{theorem}
We call $M_b$ a sample-complexity bound. However, if the training data $\Theta$ is noisy (specifically, $\mathscr{L}$ sometimes encounters flipped labels $c(\mathbf{x}) \oplus 1$), the sample-complexity bound in Eq.~(\ref{eq:M_b_no_eta}) should be modified as~\cite{Angluin1994}
\begin{eqnarray}
M_{b,\eta} = \frac{2}{\epsilon^2 \left(1 - 2\eta\right)^{2}}\ln{\left(\frac{2\abs{H}}{\delta}\right)},
\label{eq:M_b}
\end{eqnarray}
where $\eta \in \left[0, \frac{1}{2}\right)$ is the portion of the noisy pairs $(\mathbf{x}, c(\mathbf{x}) \oplus 1)$ involved in $\Theta$. However, we should note that in this case, i.e., where the noise $\eta$ exists, the sample-complexity bound $M_{b,\eta}$ is not perfectly tight; in other words, if $\abs{\Theta} \ge M_{b,\eta}$, then $\mathscr{L}$ can be ($\epsilon$, $\delta$)-PAC learner, but its inverse is not generally guaranteed.

Now let us consider a learner $\mathscr{L}$ who is accessible to a dataset $\Theta$ and we presume that $\mathscr{L}$ is a ($\epsilon$, $\delta$)-PAC learner. By recalling Eq.~(\ref{eq:M_b}) and using $\abs{\Theta} \ge M_{b,\eta}$, we can derive the following: For a finite $\epsilon$,
\begin{eqnarray}
\delta \ge e^{-\frac{1}{2} \epsilon^2 (1-2\eta)^2 \abs{\Theta}}.
\label{eq:delta}
\end{eqnarray}
Note here that we focus on the lower bound values of $\delta$ with respect to $\abs{\Theta}$, $\epsilon$, and $\eta$, and hence, the model complexity $\abs{H}$ is not considered. Then, we derive a corollary:
\begin{corollary}
Given $C$, assume that a learner $\mathscr{L}$ capable of accessing the dataset $\Theta$ attempts to be a PAC learner with $R(h,c) \le \epsilon$. Then, from {\bf Theorem~\ref{thm:1}} and Eq.~(\ref{eq:delta}), the following holds: $\mathscr{L}$ can be a ($\epsilon$, $\delta$)-PAC learner if $\mathscr{L}$ set the lower bound of $\delta$ such that
\begin{eqnarray}
\delta \ge e^{-\gamma \abs{\Theta}},
\label{eq:L(n)_bound}
\end{eqnarray}
where ${\gamma}=\frac{\epsilon^2 (1-2\eta)^2}{2}$. Here, $M_{b,\eta} \le \abs{\Theta} \le \text{poly}(\frac{1}{\epsilon}, \frac{1}{\delta}, \abs{C})$.
\label{corol:1}
\end{corollary}
This corollary represents the lower bound of $\delta$ to ensure that $\mathscr{L}$ becomes a ($\epsilon$, $\delta$)-PAC learner for a given dataset $\Theta$ and allowed learning inaccuracy $\epsilon$. However, we note that Eq.~(\ref{eq:L(n)_bound}) does not restrict the maximum learning probability achievable by $\mathscr{L}$. This is because the sample-complexity bound in Eq.~(\ref{eq:M_b}) is not tight.

\subsection{Learning probability} 

In general, a learning process involves finding a hypothesis $h \in H$ and testing whether the identified $h$ can be qualified as the true solution, in which a certain size of data within $\Theta$ is consumed. The learning accuracy $R(c,h) \le \epsilon$ can be addressed by a halting rule, which determines when the learning process is complete. In this context, we introduce the notion of learning probability, defined as follows:
\begin{definition}
For a given dataset $\Theta$ and a desired level of the learning accuracy, i.e., $R(c,h) \le \epsilon$, the learning probability, denoted as $P_L(\abs{\Theta}, \epsilon)$, is defined as the probability of completing a learning with no more than the number $\abs{\Theta}$ of training data.
\end{definition}

To understand the learning probability, we can cast a simple model based on so-called the random test model. In this model, the learning is iterated to achieve $h=c$ by consuming the data. Here, the inaccuracy $\epsilon$ is not considered because the random test is basically an exact model. Then, the learning probability at any $\abs{\Theta}$-th iteration round can be approximated as
\begin{eqnarray}
P_L^{\text{rs}}(\abs{\Theta}) = \sum_{k=1}^{\abs{\Theta}} p (1-p)^{k-1} \simeq 1 - e^{- \xi \abs{\Theta}},
\label{eq:lp_rs}
\end{eqnarray}
where $p$ is a probability of a randomly selected $h$ being the solution $c$, and $\xi^{-1}$ is the rate parameter to characterize the model performance. The probability $p$ depends only on a distribution ${\cal D}$ at each selection, and has no influence on the subsequent learning rounds. Here we assume the use of only a single sample, i.e., ($\mathbf{x}$, $c(\mathbf{x})$), for the learning, which does not affect the generalization of Eq.~(\ref{eq:lp_rs}). Here, even if an arbitrary batch of samples is used in a round of the learning, its effect can be incorporated into $\xi$. Given that the learning probability can be viewed as a cumulative distribution function, the rate parameter $\xi^{-1}$ can directly be interpreted as the average number of data consumption to achieve $c$. Such a toy model is often considered to establish a worst-case bound of the learning performance. Thus, throughout this work, our discussion focuses on learning algorithm that are at least superior to this random selection. Hence, we premise the following condition as fundamental:
\begin{eqnarray}
P_L(\abs{\Theta}, \epsilon) \ge P_L^{\text{rs}}(\abs{\Theta}),
\label{eq:rn_bound}
\end{eqnarray}
where $P_L(\abs{\Theta}, \epsilon)$ a learning probability for an {\em arbitrary} learning algorithm. 

Then, we note the following:
\begin{remark}
For given dataset $\Theta$ and fixed $\epsilon$, the learning probability $P_L(\abs{\Theta}, \epsilon)$ directly corresponds to the confidence $1-\delta$.
\label{remark:1}
\end{remark}
The immediate consequence of our note in {\bf Remark~\ref{remark:1}} is that we can establish a link between the learning probability and the PAC learning. For example, it allows us to rewrite the ($\epsilon$, $\delta$)-PAC learning condition of Eq.~(\ref{eq:def_PAC}) as
\begin{eqnarray}
P_L(\abs{\Theta},\epsilon) \ge 1-\delta.
\label{eq:revised_PAC_LP}
\end{eqnarray}
Here, the crucial point is that the theoretical statement has been transformed into practically assessable metrics. Consequently, based on Eq.~(\ref{eq:revised_PAC_LP}), the necessary $\epsilon$ and $\delta$ for a given $\mathscr{L}$ to qualify as a PAC learner can be analyzed using the accessible physical quantity, i.e., the learning probability (as shown later). For unspecified learner $\mathscr{L}$, the theoretical framework of the computational learning remains valid, allowing the PAC learnability to be specified by the size of the training data (using {\bf Theorem.~\ref{thm:1}}).

\subsection{Condition for authorized-learner attainable learning probability} 

In this subsection, we establish a condition that permit or restrict the qualitative differences in the learning outcomes achievable by an authorized learner, say $\mathscr{L}_A$, and an eavesdropping learner, say $\mathscr{L}_E$. Here, the term ``authorized'' means that the learner has the permission to access the training data. To this end, we first consider the use of a sort of ``classical-quantum hybrid'' encoded data as~\cite{Lee2019,Song2021b}:
\begin{eqnarray}
\Theta_Q=\{ (\mathbf{x}, \ket{c(\mathbf{x})}) | \mathbf{x} \in {\cal X}, c \in C \}.
\label{eq:Q_data}
\end{eqnarray}
Comparing with the use of $\Theta$ in Eq.~(\ref{eq:l_data}), the notable point is that the label $c(\mathbf{x})$ is encoded into a qubit, which is called ``quantum label encoding.'' Then, let us consider the following scenario: $\mathscr{L}_A$ accesses to, say, a data center, via a classical and quantum channel, denoted as ${\cal C}_C$ and ${\cal C}_Q$, respectively. The concept of the data center is often casted, where it usually possesses the (big) data for learning~\cite{Liu2023,Liu2024}. Here, a quantum protocol ${\cal P}$ can be employed to transmit $\Theta_Q$ to $\mathscr{L}_A$. In such a setting, $\mathscr{L}_E$ can intrude (${\cal C}_C$, ${\cal C}_Q$) to attain his/her own data $\Xi_{Q,E} = \{ (\mathbf{x}, \hat{\rho}_E(\mathbf{x})) \}$, where $\hat{\rho}_E(\mathbf{x})$ is the state of labels. We indicate that $\hat{\rho}_E(\mathbf{x})$ is not pure, i.e., $\hat{\rho}_E(\mathbf{x}) \neq \ket{c(\mathbf{x})}\bra{c(\mathbf{x})}$, due to the noise
\begin{eqnarray}
\eta_E = 1 - \max_{{\cal S}_E({\cal P})} \overline{\cal F}_E,
\label{eq:etaE}
\end{eqnarray}
where ${\cal S}_E({\cal P})$ denotes $\mathscr{L}_E$'s eavesdropping strategy in the protocol ${\cal P}$ and $\overline{\cal F}_E$ is the ensemble fidelity, given by
\begin{eqnarray}
\overline{\cal F}_E = \frac{1}{\abs{\Xi_{Q,E}}}\sum_{\mathbf{x} \in \Xi_{Q,E}} \text{Tr}{ \left( \hat{\rho}_E(\mathbf{x}) \ket{c(\mathbf{x})}\bra{c(\mathbf{x})} \right)}.
\end{eqnarray}
Here, we note that, due to the principle of quantum mechanics, $\mathscr{L}_E$'s intervention inevitably causes the disturbance, i.e., noise, in ${\cal C}_Q$~\cite{Fuchs1996,Fuchs1997}. In consequence, $\mathscr{L}_A$ would also have an imperfect dataset $\Xi_{Q,A} = \{ (\mathbf{x}, \hat{\rho}_A(\mathbf{x})) \}$ with the noise $\eta_A$. This noise factor $\eta_A$ could be understood as the ``quantum-bit-error rate'' in the context of quantum-key-distribution scenario~\cite{Bocquet2011}. Here, it is generally assumed that
\begin{eqnarray}
\abs{\Xi_{Q,E}} \le \abs{\Xi_{Q,A}} \le \abs{\Theta_Q}.
\label{eq:data_sizes}
\end{eqnarray} 

On the basis of the previous discussions, we present a proposition:
\begin{proposition}
There is a robust protocol, denoted as ${\cal P}$, which restricts any $\mathscr{L}$'s strategy ${\cal S}_E({\cal P})$ to satisfy
\begin{eqnarray}
\left( \eta_A < \eta^{\star} \right) \land \left( \eta_E < \eta^{\star} \right).
\label{eq:eta_critical}
\end{eqnarray}
Here, $\eta^{\star}$ represents a critical threshold where the reductions in $\eta_A$ and $\eta_E$ coincide. The determination of $\eta^{\star}$ hinges on the strategies ${\cal S}_E({\cal P})$ formulated within the framework of quantum mechanics.
\label{prop:1}
\end{proposition}
Here, we have assumed the following: [{\bf A.1}] Firstly, $\mathscr{L}_E$ has no influence on the choices of $\mathbf{x}$ to set $\Theta$. [{\bf A.2}] Second is that $\mathscr{L}_E$ cannot intrude the devices of data center and $\mathscr{L}_A$ directly; e.g., the random number generator and/or measurements. [{\bf A.3}] Lastly, $\mathscr{L}_E$'s strategy has to obey the quantum mechanics. 

Now, let us consider that $\mathscr{L}_A$ cast a robust protocol ${\cal P}$ satisfying Eq.~(\ref{eq:eta_critical}). As indicated, during the data transmission, the prepared datasets $\Xi_{Q,A}$ and $\Xi_{Q,E}$ by $\mathscr{L}_A$ and $\mathscr{L}_E$ are noisy. In this situation, the following holds (from {\bf Corollary~\ref{corol:1}}): For the given $\Xi_{Q,A}$ and $\Xi_{Q,E}$, [{\bf C.1}] $\mathscr{L}_A$ is ensured to be ($\epsilon_A$, $\delta_A$)-PAC learner by limiting $\delta_A \ge \delta_A^{\star}$ and [{\bf C.2}] $\mathscr{L}_E$ is ensured to be ($\epsilon_E$, $\delta_E$)-PAC learner by limiting $\delta_E \ge \delta_E^{\star}$. Here, $\delta_A^{\star}$ and $\delta_E^{\star}$ are defined as
\begin{eqnarray}
\delta_A^{\star} = e^{-\gamma_A \abs{\Xi_{Q,A}}}~\text{and}~\delta_E^{\star} = e^{-\gamma_E \abs{\Xi_{Q,E}}},
\label{eq:delta_star}
\end{eqnarray}
where $\gamma_{A}=\frac{\epsilon_{A}^2 (1-2\eta_{A})^2}{2}$ and $\gamma_{E}=\frac{\epsilon_{E}^2 (1-2\eta_{E})^2}{2}$. Then, we can obtain the following:
\begin{theorem}
For a protocol ${\cal P}$ satisfying Eq.~(\ref{eq:eta_critical}), $\mathscr{L}_A$ is ensured to be a ($\epsilon_A$, $\delta_A$)-PAC learner by limiting $\delta_A \ge \delta_A^{\star}$. Here, if $\eta_A > \eta^{\star}$ is secured from $\Xi_{Q,A}$, there is no condition that ensures $\mathscr{L}_E$ becomes a ($\epsilon_E$, $\delta_E$)-PAC learner satisfying
\begin{eqnarray}
\left( \epsilon_E \le \epsilon_A \right) \land \left( \delta_E \le \delta_A \right).
\label{eq:l_qual_e_d}
\end{eqnarray}
\label{thm:2}
\end{theorem}
The proof of this theorem is straightforward. At first, let $\epsilon_E = \epsilon_A$. Then, from Eq.~(\ref{eq:data_sizes}) and Eq.~(\ref{eq:eta_critical}), we can prove that
\begin{eqnarray}
\eta_A < \eta^{\star} \Rightarrow \delta_A^{\star} < \delta_E^{\star}.
\label{eq:eta_delta}
\end{eqnarray}
Thus, even if $\mathscr{L}_E$ is ensured to be a PAC learner exhibiting $\epsilon_E = \epsilon_A$ (as per [{\bf C.2}]), the condition $\delta_E \le \delta_A$ cannot be satisfied. On the other hand, $\mathscr{L}_E$ can consider the setting $\delta_A^{\star} = \delta_E^{\star}$. However, this is only possible when $\epsilon_E > \epsilon_A$ [see Eq.~(\ref{eq:delta_star})]. Consequently, {\bf Theorem~\ref{thm:2}} holds. Here, it should be noted that {\bf Theorem~\ref{thm:2}} does not, in principle, prohibit $\mathscr{L}_E$ from returning a single learning outcome that satisfies Eq.~(\ref{eq:l_qual_e_d}); however, does not ensure it. If the tight bound of Eq.~(\ref{eq:M_b}) is derived, {\bf Theorem~\ref{thm:2}} can be extended to a stronger condition, namely that prohibits $\mathscr{L}_E$ from achieving Eq.~(\ref{eq:l_qual_e_d}) in any cases.

Our {\bf Theorem~\ref{thm:2}} has noteworthy implications: (i) Firstly, it establishes a theoretical condition, based on the quantum theory, that limits the guarantee of learning quality achievable by any eavesdropping learner. The validity of {\bf Proposition~\ref{prop:1}}, which underpins {\bf Theorem~\ref{thm:2}}, is grounded in the quantum no-cloning theorem~\cite{Scarani2005,Dang2007} and/or the tradeoff between information gain and disturbance~\cite{Fuchs1996,Banaszek2001}. (ii) Secondly, an authorized learner $\mathscr{L}_A$ can confirm the limitations imposed on $\mathscr{L}_E$ based, solely, on the noise degree in his/her own dataset $\Xi_{Q,A}$. (iii) Lastly, the validity of {\bf Theorem~\ref{thm:2}} is contingent upon the existence of the transmission protocol ${\cal P}$ that meets the criteria in Eq.~(\ref{eq:eta_critical}). Thus, by minimizing $\eta^\star$ by developing more efficient quantum encoding schemes, one can effectively lower the learning quality ensured for $\mathscr{L}_E$.

\section{CNN-based image classification}\label{sec:3}

The theoretical proofs demonstrating the learning outcome superiority of $\mathscr{L}_A$ over $\mathscr{L}_E$ may not always be feasible. For instance, the learning models required to observe the superiority of $\mathscr{L}_A$'s learning (as detailed in {\bf Theorem~\ref{thm:2}}) might not be available in real-world scenarios, or the amount of data $\abs{\Theta_Q}$ required might be excessively large. Therefore, in this section, we aim to validate the theoretical insights established in the previous Sec.~\ref{sec:2} by using the convolutional neural networks (CNNs) for image classification.

\subsection{Protocol} 

\begin{figure}[t]
	\centering
	\includegraphics[width=0.60\textwidth]{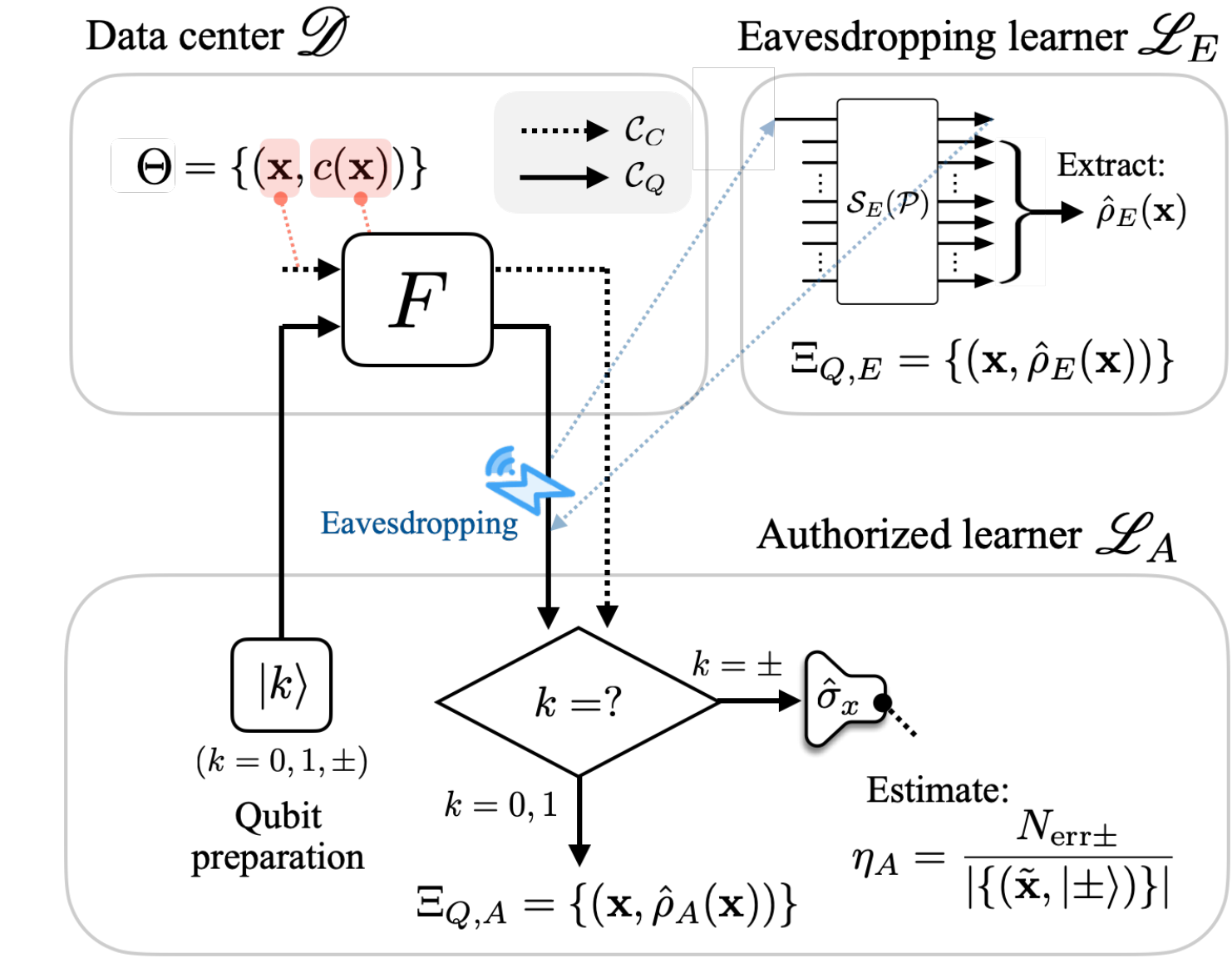}
	\caption{A schematic of the protocol ${\cal P}$. $\mathscr{L}_A$ prepares a qubit state $\ket{k}$ ($k=0,1,\pm$) and send it to $\mathscr{D}$ via ${\cal C}_Q$. The state $\ket{k}$ is passed through a function $F$ with a chosen $\mathbf{x}$. $F$ generate a pair $(\mathbf{x}, \ket{c(\mathbf{x}) \oplus k})$ when $k=0,1$, and $(\mathbf{x}, \ket{k})$ when $k=\pm$. The output pair is returned to $\mathscr{L}_A$ via ${\cal C}_C$ and ${\cal C}_Q$. At this point, $\mathscr{L}_A$ obtains a training data for $k=0,1$. For $k=\pm$, $\mathscr{L}_A$ performs a $\hat{\sigma}_x$ measurement to check if the incoming qubit has been disturbed by $\mathscr{L}_E$. By repeating this process, $\mathscr{L}_A$ obtains a noisy dataset $\Xi_{Q,A} = \{ (\mathbf{x}, \hat{\rho}_A(\mathbf{x})) \}$, estimating $\eta_A$ [as in Eq.~(\ref{eq:eta_A})]. Following the strategy ${\cal S}_E({\cal P})$, $\mathscr{L}_E$ also obtains a dataset $\Xi_{Q,E} = \{ (\mathbf{x}, \hat{\rho}_E(\mathbf{x})) \}$ with $\eta_E$.}
	\label{fig:protocol}
\end{figure}

In this subsection, we introduce a data transmission protocol ${\cal P}$ described in {\bf Proposition~\ref{prop:1}}, which is a slightly modified version of the method from Ref.~\cite{Song2021a}. Firstly, let us consider a data center, denoted as $\mathscr{D}$, which has a finite set of the training data $\Theta = \{ (\mathbf{x}, c(\mathbf{x})) \}$ (where $\abs{\Theta} \ll 2^n$). Here, assume that the inputs $\mathbf{x}$ can be shared publicly\footnote{This assumption is not arbitrary. In fact, in the encoding in Eq.~(\ref{eq:Q_data}), the input values $\mathbf{x}$ are treated as classical. In $\mathscr{L}_E$'s strategy, which includes quantum theory, $\mathbf{x}$ can be cloned or intercepted.}, but the labels $c(\mathbf{x}) \in \{0, 1\}$ for each input $\mathbf{x}$ should be provided only to $\mathscr{L}_A$. In such a setting, the protocol runs as follows: $\mathscr{L}_A$ prepares a state $\ket{k}$ from the set $\{ \ket{0}, \ket{1}, \ket{\pm} \}$ at random (i.e., with $\frac{1}{4}$ probability) and sends it to $\mathscr{D}$ through ${\cal C}_Q$. For the incoming $\ket{k}$, $\mathscr{D}$ chooses an input $\mathbf{x}$ involved in $T_\mathscr{D}$. Then, the pair $(\mathbf{x}, \ket{k})$ ($k=0,1,\pm$) is processed by a function $F$, which is equipped in $\mathscr{D}$. The function $F$ transforms the qubit state such that $\ket{k}$ becomes $\ket{c(\mathbf{x}) \oplus k}$ for $k = 0,1$; the state $\ket{\pm}$ remains unchanged for $k = \pm$\footnote{The function $F$, sometimes referred to as an ``oracle,'' provides the direct access to the information of $c$. While this $F$ is typically treated as a black-box operation, it can be implemented in a classical-quantum hybrid circuit level. For details about the realization of $F$, see Appendix~A in Ref.~\cite{Song2021a} or Ref.~\cite{Song2021b}.}. The encoded pairs $(\mathbf{x}, \ket{c(\mathbf{x}) \oplus k})$ or $(\mathbf{x}, \ket{\pm})$ passed through $F$ is delivered to $\mathscr{L}_A$ via ${\cal C}_Q$ and ${\cal C}_C$. Thus, $\mathscr{L}_A$ obtains a pair $(\mathbf{x}, \ket{c(\mathbf{x}) \oplus k})$ for the learning when $k=0,1$, and $(\mathbf{x}, \ket{k})$ when $k=\pm$. Here, for $k=\pm$, $\mathscr{L}_A$ performs $\hat{\sigma}_x$ measurement to check whether the state $\ket{\pm}$ has been disturbed in ${\cal C}_Q$, thereby detecting any eavesdropping learners. By repeating this process, $\mathscr{L}_A$ is allowed to obtain a (noisy) dataset $\Xi_{Q,A}=\{ (\mathbf{x}, \hat{\rho}_A(\mathbf{x})) \}$. If there is no $\mathscr{L}_E$'s eavesdropping, $\mathscr{L}_A$ can get a clean training data $\Theta_Q = \{ (\mathbf{x}, \ket{c(\mathbf{x}) \oplus k}) \}$ for $k=0,1$. The noise factor $\eta_A$ can be estimated by counting the unexpected changes, i.e., $\ket{\pm} \to \ket{\mp}$, such that
 \begin{eqnarray}
 \eta_A = \frac{N_{\text{err}\pm}}{\abs{\{ (\tilde{\mathbf{x}}, \ket{\pm}) \}}},
 \label{eq:eta_A}
 \end{eqnarray}
 where $N_{\text{err}\pm}$ denotes the number of changed results in $\mathscr{L}_A$'s $\hat{\sigma}_x$ measurements and $\{ (\tilde{\mathbf{x}}, \ket{\pm}) \}$ is the set of the invalid training data. Here, the tilde symbol above $\mathbf{x}$ is used to distinguish it from those in $\Xi_{Q,A}$; namely, $\{ \mathbf{x} \} \cap \{ \tilde{\mathbf{x}} \}$ is null, and $(\tilde{\mathbf{x}}, \hat{\rho}(\tilde{\mathbf{x}})) \notin \Xi_{Q,A}$. The eavesdropping learner $\mathscr{L}_E$, faithfully following the eavesdropping strategy ${\cal S}_E({\cal P})$, can also extract a dataset $\Xi_{Q,E} = \{ (\mathbf{x}, \hat{\rho}_E(\mathbf{x})) \}$ with $\eta_E$. We note that the value of $\eta_E$ cannot be estimated in ${\cal S}_E({\cal P})$. A schematic of this protocol ${\cal P}$ is depicted in Fig.~\ref{fig:protocol}.

The protocol ${\cal P}$ described above allows us to identify a threshold value $\eta^{\star}$ satisfying Eq.~(\ref{eq:eta_critical}). For this, we can cast a useful quantity: the information accessible to $\mathscr{L}_A$ (or $\mathscr{L}_E$) from the dataset $\Theta$, represented by the mutual information $I_{\Theta A}$ (or $I_{\Theta E}$)~\cite{Devetak2005,Cai2004}. Noting that $I_{\Theta E}$ can be maximized by the eavesdropping strategies ${\cal S}_E({\cal P})$, let us consider the following quantity:
\begin{eqnarray}
I_{\Theta A} - \max_{{\cal S}_E({\cal P})} I_{\Theta E},
\label{eq:mutual_inf}
\end{eqnarray}
which can quantify the threat of $\mathscr{L}_E$'s eavesdropping on the training data. Here, we provide a conjecture 
\begin{conjecture}
A higher quality learning outcome for $\mathscr{L}_A$ is guaranteed when Eq.~(\ref{eq:mutual_inf}) has a positive value, specifically, when the following condition, known as Holevo's condition~\cite{Holevo2011book}, is met:
\begin{eqnarray}
I_{\Theta A} \ge \max_{{\cal S}_E({\cal P})} I_{\Theta E}.
\label{eq:Holevo_c}
\end{eqnarray}
\label{conj:1}
\end{conjecture}
Given that the learning is fundamentally a data-driven process, this conjecture, which indicates the direct correlation between the amount of information extractable from training data and learning outcome quality, is intuitive~\cite{Schumacher1997,Chen2016}. 
 
 As explained in Sec.~\ref{sec:2}-C, the protocol ${\cal P}$, based on quantum theory, reflects the tradeoff between the quality of the datasets obtained by $\mathscr{L}_A$ and $\mathscr{L}_E$, specifically, between $\eta_A$ and $\eta_E$. This leads to a tradeoff between $I_{\Theta A}$ and $\max_{{\cal S}_E({\cal P})} I_{\Theta E}$, with the threshold value $\eta^{\star}$ being identified at the point where $I_{\Theta A}$ and $\max_{{\cal S}_E({\cal P})} I_{\Theta E}$ are equal. Here, we note that as the strategy ${\cal S}_E({\cal P})$ of $\mathscr{L}_E$ improves, the value in Eq.~(\ref{eq:mutual_inf}) decreases to zero, and $\eta^{\star}$ becomes smaller. This implies that better ${\cal S}_E({\cal P})$ requires stricter condition to ensure superior learning outcome for $\mathscr{L}_A$. Currently, the best strategy ${\cal S}_E({\cal P})$, so-called the collective attacks, gives $\eta^{\star} \simeq 0.11$. For ${\cal S}_E({\cal P})$ corresponding to memoryless and individual attacks, we can identify $\eta^{\star} \simeq 0.154$ and $\eta^{\star} \simeq 0.146$, respectively. Such the discussions are deeply rooted in the quantum secure communication scenario (for more details, see Ref.~\cite{Bocquet2011}).

\subsection{Image classification} 

\begin{figure}[t]
	\centering
	\includegraphics[width=1.00\textwidth]{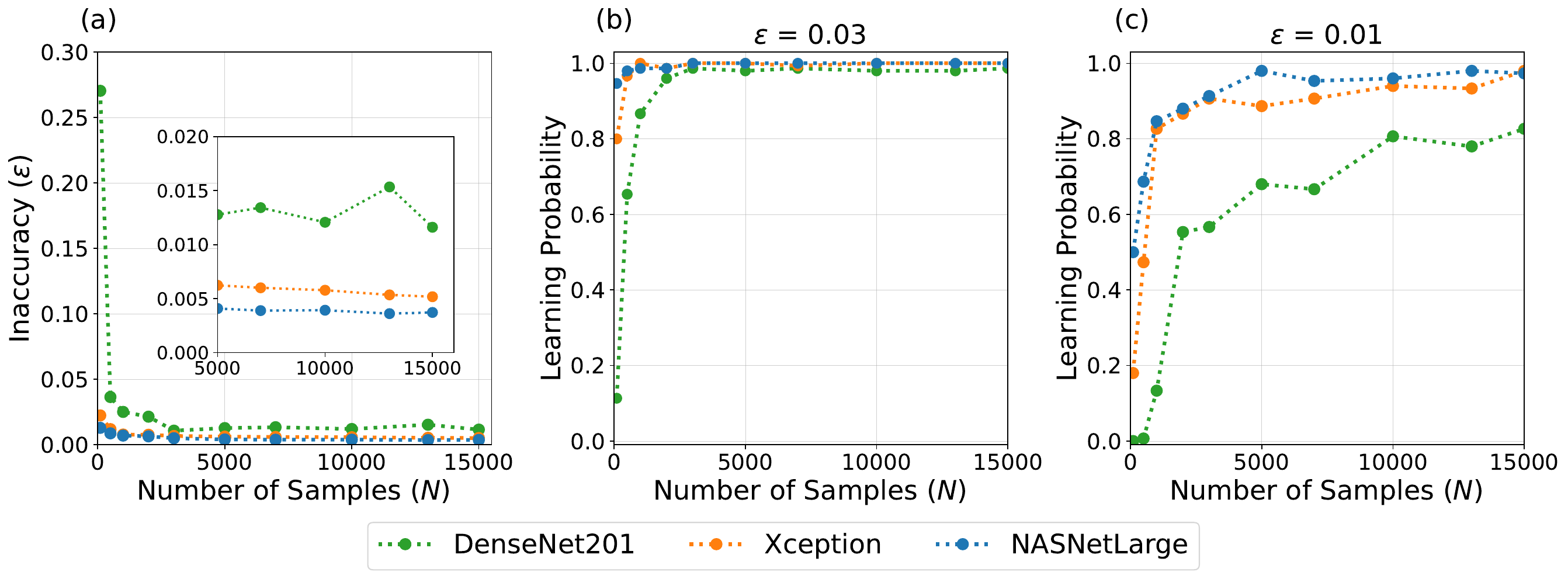}
	\caption{(a) First, we plot the graphs showing the average accuracy obtained to the size of the training data $\abs{\Theta_Q}$ used for each CNN model---DN, XC, and NNL. Generally, a larger size of $\abs{\Theta_Q}$ allows for more accurate learning, but improvements reach a plateau beyond a certain threshold depending on the used CNN model. The achievable accuracy ranks in order: NNL, XC, and DN. We run each of the three CNN models through $150$ trials of the learning and, for (a) $\epsilon_T = 0.03$ and (c) $\epsilon_T = 0.01$, we plot the learning probability $P_L(\abs{\Theta_Q}, \epsilon_T)$ based on the cumulative distribution of the used data. As mentioned earlier, these learning probability graphs provide insights into whether $\mathscr{L}_A$ qualifies as a PAC learner.}
	\label{fig:no_LE1}
\end{figure}

Here we investigate whether the theoretically established aspects of the learning superiority for $\mathscr{L}_A$ appear in practice. We thus design the task as follows: $\mathscr{L}_A$ receives the training dataset from $\mathscr{D}$ via ${\cal P}$, and subsequently, performs the classification learning by employing the convolutional neural networks (CNNs). For multifaceted analysis, we employ three different types of the pre-trained models: DenseNet201 (DN), Xception (XC), and NASNetLarge (NNL)\footnote{DenseNet201 is a simplified model optimized for training speed, making it useful for fundamental analyses, such as learning feasibility. NASNetLarge is a more complex model focused on optimizing the quality of the learning outcomes. Xception is considered a balanced model that incorporates characteristics of both aforementioned models [See Fig.~\ref{fig:no_LE1}(a)-(c)].}. We use an image-set consisting of cats and dogs provided by ImageNet\footnote{The official website of ImageNet is available at http://www.image-net.org.}. Each image is represented as an input $\mathbf{x}$ and the corresponding label is encoded as $\ket{c(\mathbf{x})=\text{``cat''}}$ (say, $\ket{0}$) or $\ket{c(\mathbf{x})=\text{``dog''}}$ (say, $\ket{1}$). The image data $\Theta_Q = \left\{ (\mathbf{x}_i, \ket{c(\mathbf{x}_i)}) \right\}$ is then transmitted to $\mathscr{L}_A$ via ${\cal P}$. The test dataset ($\not\subseteq \Theta_Q$) used to evaluate $R(h,c)$ is assumed to be fully classical. This is to ensure that the learning performances of both $\mathscr{L}_A$ and $\mathscr{L}_E$ are fairly evaluated.

\begin{figure}[t]
	\centering
	\includegraphics[width=1.00\textwidth]{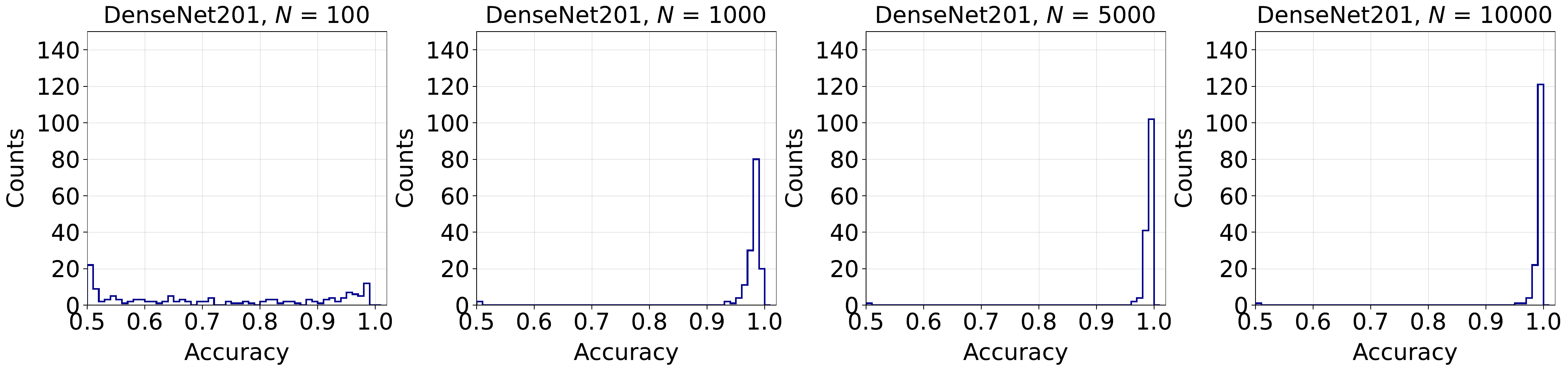}
	\includegraphics[width=1.00\textwidth]{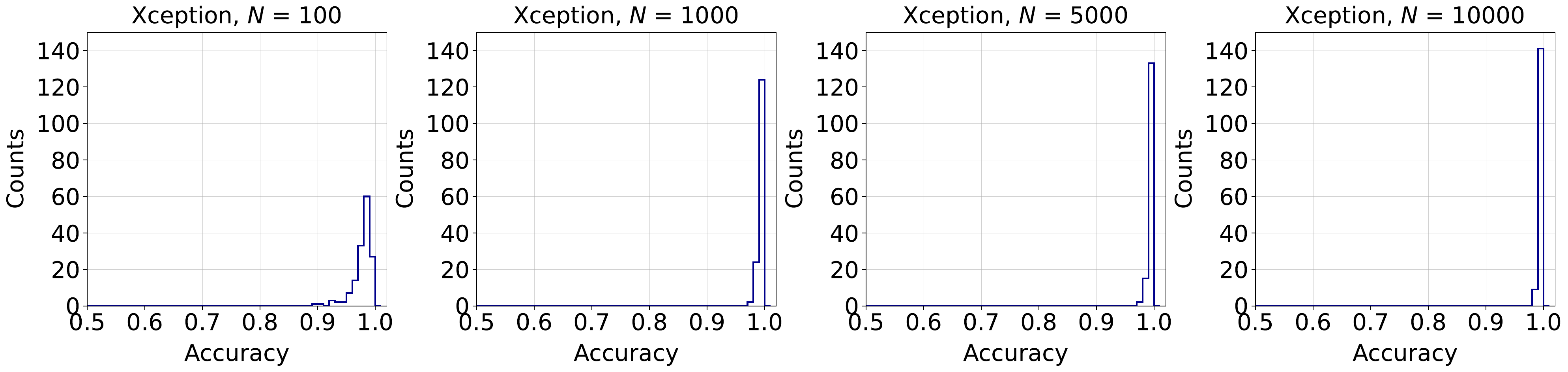}
	\includegraphics[width=1.00\textwidth]{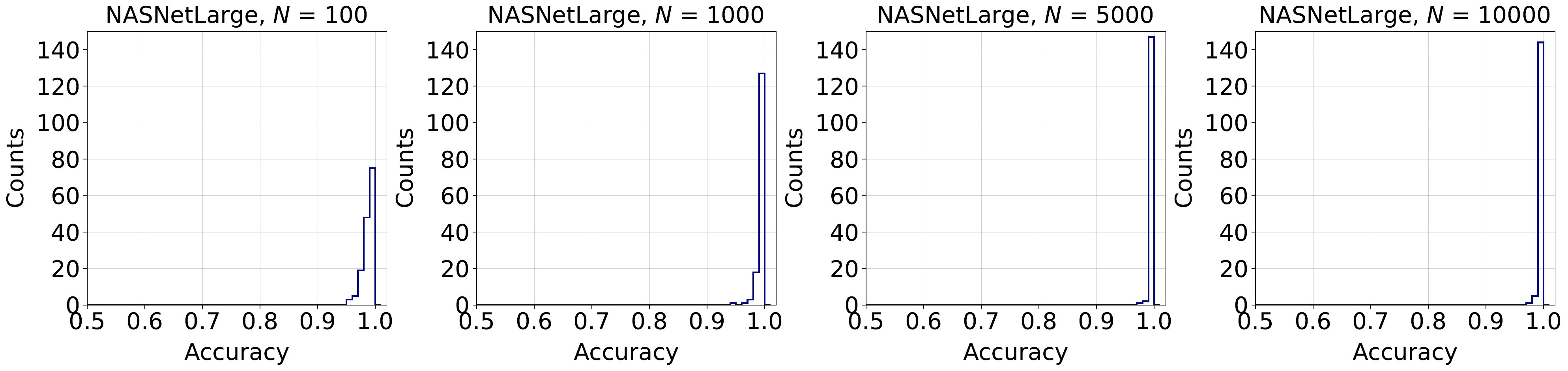}		
	\caption{Histogram distributions of the learning accuracy ($1-\epsilon$) achieved for each model (DN, XC, and NNL) and available data sizes ($\abs{\Theta_Q}=100$, $1000$, $5000$, and $10000$), with intervals of $0.01$. Generally, as the size of the available dataset increases for all models, higher learning accuracy is achieved consistently, allowing for successful completion of the CNN learning. However, with insufficient training data, the distribution of the resulting learning accuracy becomes wider, making it difficult to consistently achieve high-quality learning outcomes. This pattern is prominent in inefficient learning models.}
	\label{fig:no_LE2}
\end{figure}

{\em No eavesdropping.}---At first, assuming no eavesdropping learner(s) $\mathscr{L}_E$ and no disturbance on ${\cal P}$, we perform the numerical simulations to quantify the learning probability of $\mathscr{L}_A$. To run CNN learning simulations, we use a workstation using RTX 3080 GPUs. We begin by setting a target learning accuracy $R(h,c) < \epsilon_T$, and count the amount of training data $\subseteq \abs{\Theta_Q}$ required to achieve $\epsilon_T$. Throughout the learning process, we evaluate $R(h,c)$; if $R(h,c)$ falls below the threshold $\epsilon_T$, the learning is considered complete. For the chosen $\epsilon_T$, we obtain the learning probability $P_L(\abs{\Theta_Q}, \epsilon_T)$ by repeating the learning. The learning outcomes from $150$ independent runs in each pre-trained model---DN, XC, and NNL---are analyzed. The results are summarized in Fig.~\ref{fig:no_LE1} and~\ref{fig:no_LE2}. Firstly, we give the achievable $\epsilon$ for different data sizes $\abs{\Theta_Q}$ in Fig.~\ref{fig:no_LE1}(a). The learning accuracy improves with data size $\abs{\Theta_Q}$ but reaches a threshold. This threshold depends on the used CNN model. We then plot the learning probabilities $P_L(\abs{\Theta_Q},\epsilon_T)$ for $\epsilon_T = 0.01$ and $\epsilon_T = 0.03$ in Fig.~\ref{fig:no_LE1}(b) and~\ref{fig:no_LE1}(c), respectively. As noted in {\bf Remark~\ref{remark:1}}, $P_L(\abs{\Theta_Q},\epsilon_T)$ is a measurable physical quantity that can be used to identify the confidence $1 - \delta$, allowing to define $\mathscr{L}_A$ as a ($\epsilon_T$, $\delta$)-PAC learner based on Eq.~(\ref{eq:revised_PAC_LP}). For instance, if more than $1000$ data samples are available, $\mathscr{L}_A$ using DN model can achieve an accuracy of, at least, $\epsilon = 0.03$ with the confidence over $80\%$ (i.e., $\delta \le 0.2$); in other words, $\mathscr{L}_A$ can serve as a ($\epsilon=0.03$, $\delta \le 0.2$)-PAC learner with DN when $\abs{\Theta_Q} \ge 1000$ data samples are used (see Fig.~\ref{fig:no_LE1}(b)). However, when $\epsilon_T$ is set to be $0.01$, $\mathscr{L}_A$, using DN has to set a lower confidence level to qualify as a PAC learner, as shown in Fig.~\ref{fig:no_LE1}(c). Fig.~\ref{fig:no_LE2} presents the histograms, showing the distribution of $\epsilon$ obtained by each CNN model for $\abs{\Theta_Q}=100$, $1000$, $5000$, and $10000$. The histograms show that despite performance differences, each CNN model learns reliably. However, when the training data size is insufficient, it would be challenging to achieve high learning accuracy (see Fig.~\ref{fig:no_LE2}(a) for an extreme case).

\begin{figure}[t]
	\centering
	\includegraphics[width=1.00\textwidth]{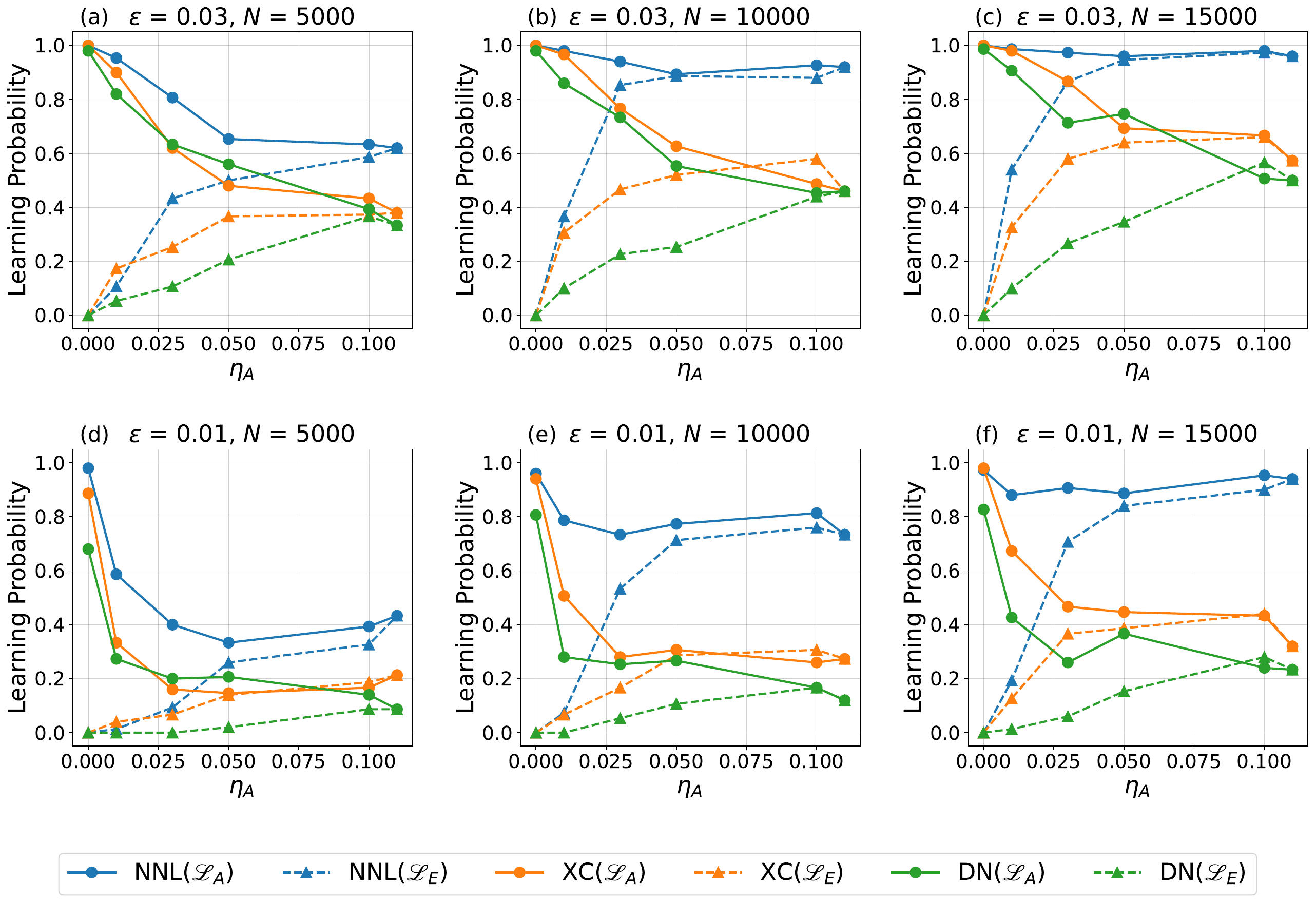}
	\caption{We plot graphs of the learning probability $P_L(\abs{\Theta_Q}, \epsilon_T)$ of $\mathscr{L}_A$ and $\mathscr{L}_E$ versus the noise level $\eta$ for each pre-trained CNN model (DN, XC, and NNL). These graphs illustrate six cases (a)-(f), with target accuracy set to $\epsilon_T = 0.03$ and $0.01$, and data sizes $\abs{\Theta_Q}$ of $5000$, $10000$, and $15000$. As predicted by {\bf Theorem~\ref{thm:2}} and {\bf Conjecture~\ref{conj:1}} in the theoretical analysis, a clear trade-off relation is observed between the quality of the learning outcomes for $\mathscr{L}_A$ and $\mathscr{L}_E$. For details, see the main text.}
	\label{fig:learning_P}
\end{figure}

\begin{figure}[t]
	\centering
	\includegraphics[width=1.0\textwidth]{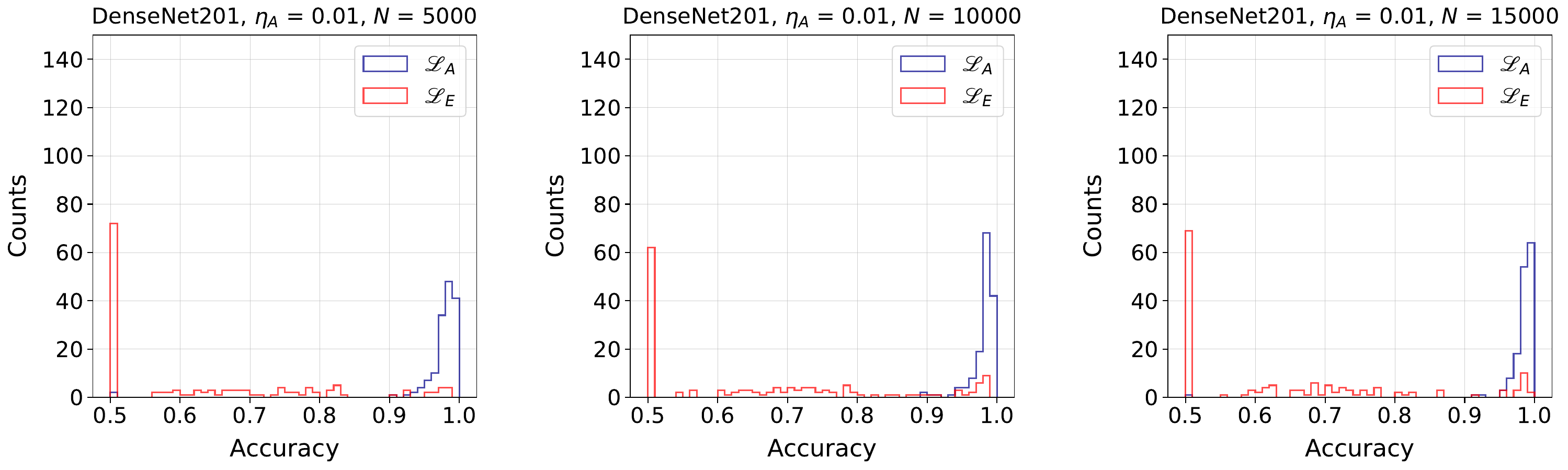}
	\includegraphics[width=1.0\textwidth]{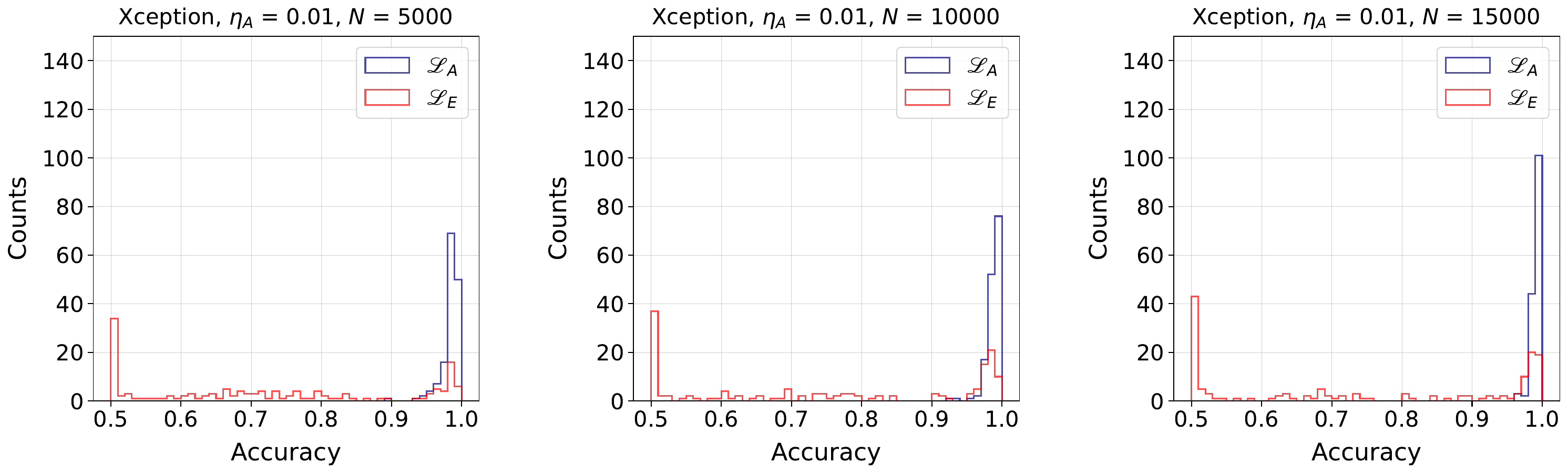}
	\includegraphics[width=1.0\textwidth]{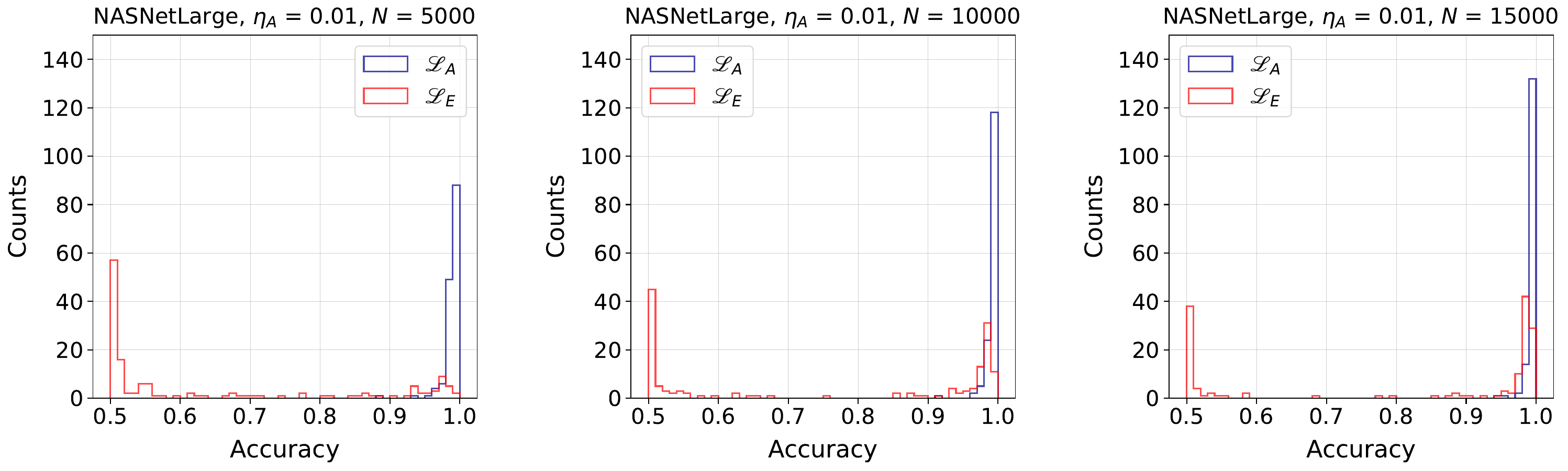}
	\caption{For a noise level of $\eta = 0.01$, i.e., when $\mathscr{L}_E$'s data extraction from $\Theta_Q$ is less aggressive to avoid detection, the histogram distributions of the learning accuracy ($1-\epsilon$) obtained by $\mathscr{L}_A$ and $\mathscr{L}_E$ through each pre-trained model (DN, XC, and NNL) are shown at intervals of $0.01$ for the data sizes $\abs{\Theta_Q}$ of $5000$, $10000$, and $15000$, respectively. In all pre-trained models, a clear difference in the quality of the learning outcomes between $\mathscr{L}_A$ and $\mathscr{L}_E$ is observed.}
	\label{fig:histogram_AE001}
\end{figure}

\begin{figure}[t]
	\centering
	\includegraphics[width=1.0\textwidth]{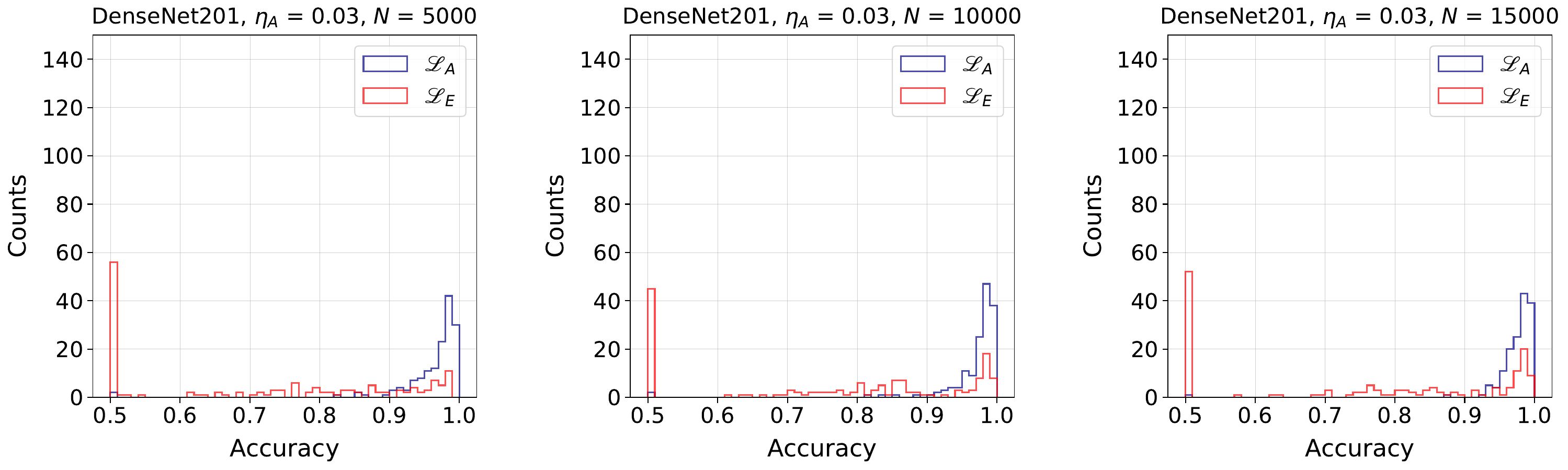}
	\includegraphics[width=1.0\textwidth]{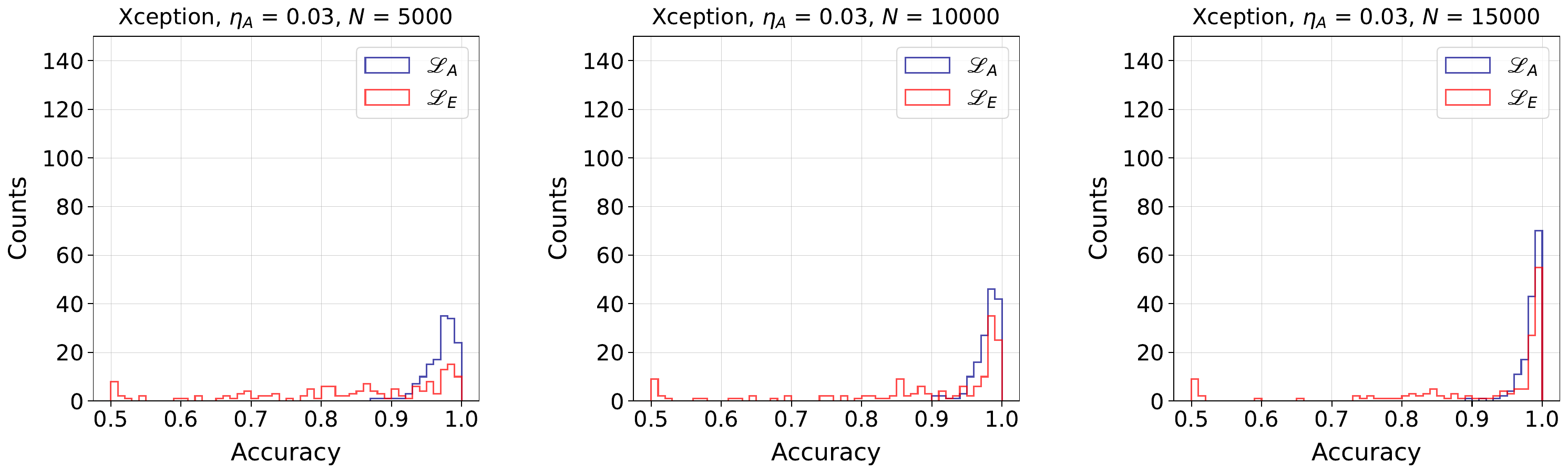}
	\includegraphics[width=1.0\textwidth]{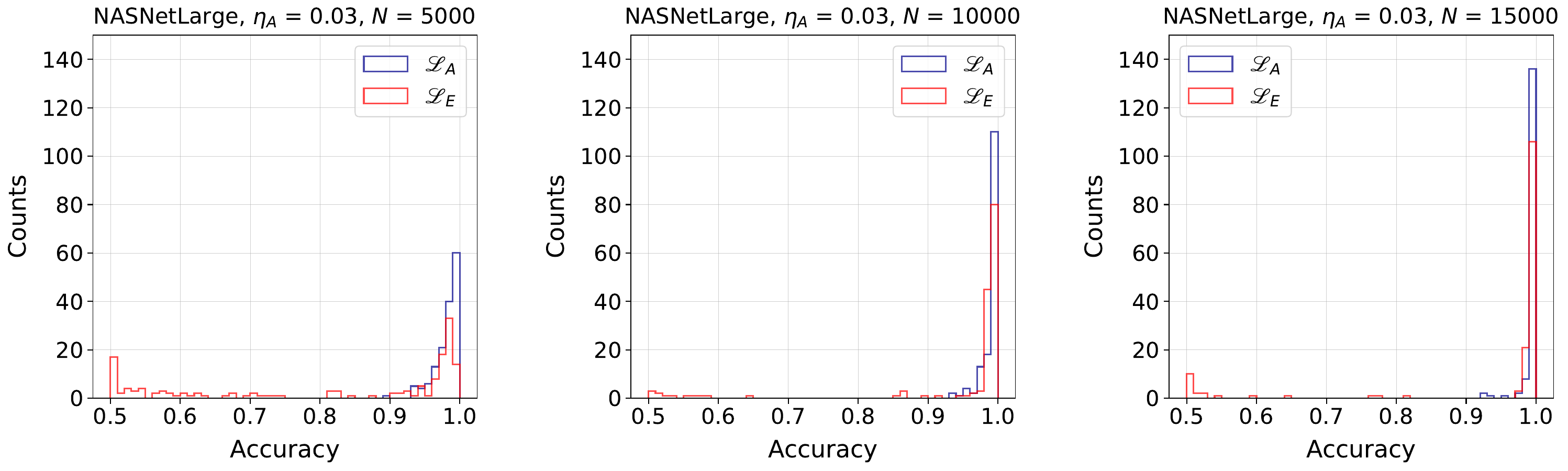}
	\caption{For a noise level of $\eta = 0.03$, indicating a more aggressive eavesdropping by $\mathscr{L}_E$, the histogram distributions of the learning accuracy ($1-\epsilon$) obtained by $\mathscr{L}_A$ and $\mathscr{L}_E$ through each pre-trained model (DN, XC, and NNL) are shown at intervals of $0.01$ for the data sizes $\abs{\Theta_Q}$ of $5000$, $10000$, and $15000$, respectively. While the probability of detecting $\mathscr{L}_E$ increases, the gap between $\mathscr{L}_A$'s and $\mathscr{L}_E$'s learning patterns begins to narrow (e.g., in XC and NNL).}
	\label{fig:histogram_AE003}
\end{figure}

\begin{figure}[t]
	\centering
	\includegraphics[width=1.0\textwidth]{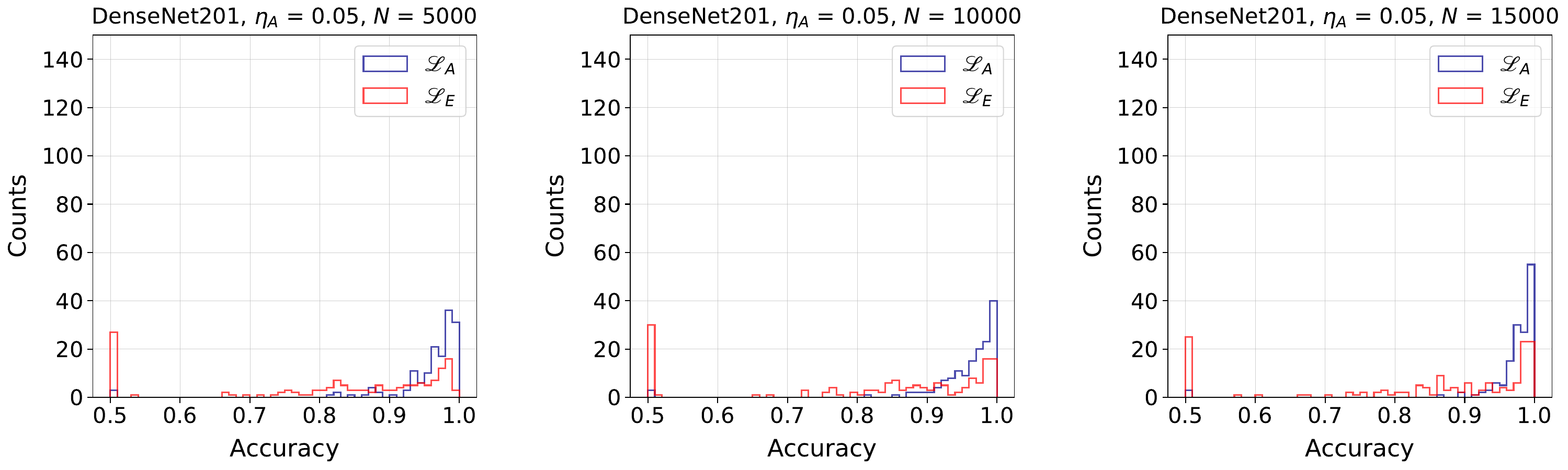}
	\includegraphics[width=1.0\textwidth]{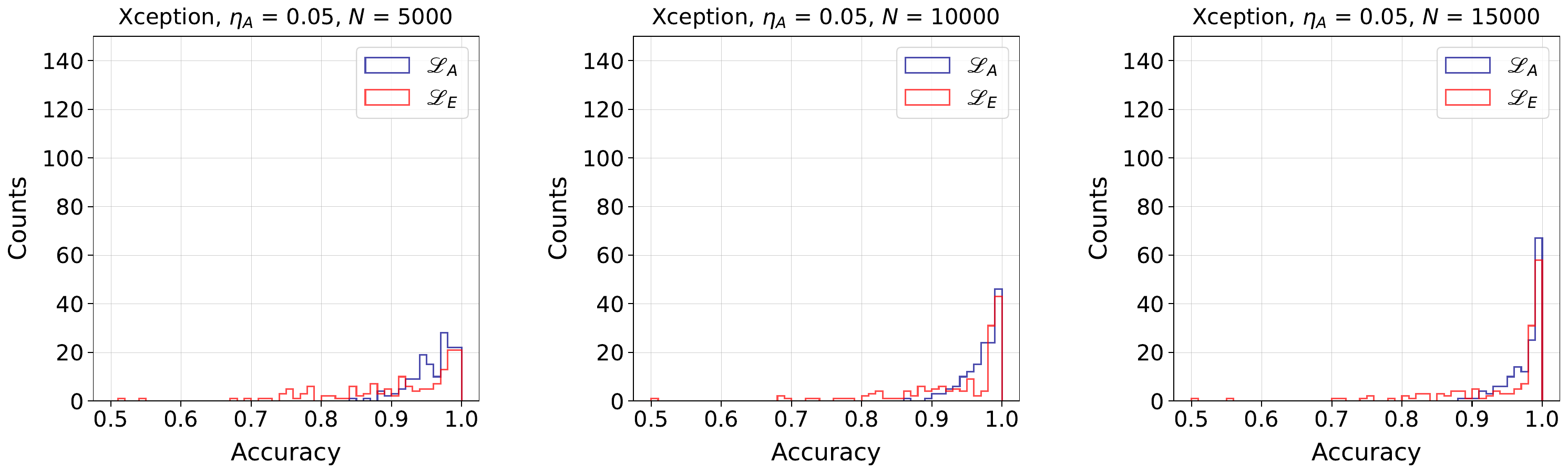}
	\includegraphics[width=1.0\textwidth]{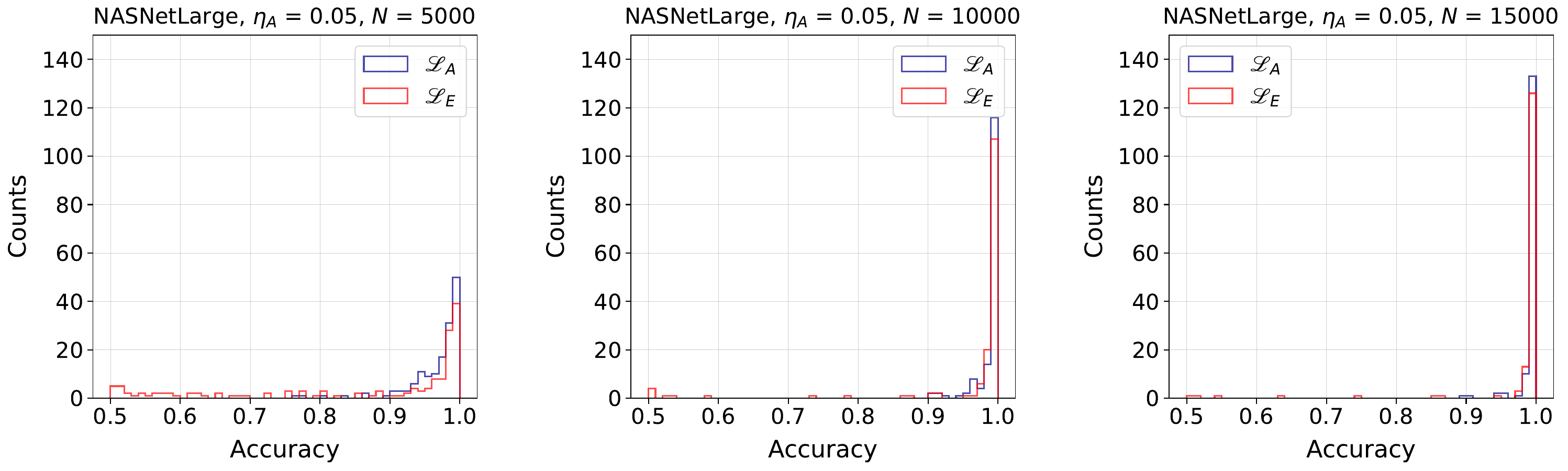}
	\caption{For a noise level of $\eta = 0.05$ (i.e., when $\mathscr{L}_E$'s data extraction from $\Theta_Q$ is relatively large), the histograms for $\mathscr{L}_A$'s and $\mathscr{L}_E$'s CNN learnings are shown. In the performance-optimized model, such as NNL, with ample training data, both $\mathscr{L}_A$ and $\mathscr{L}_E$ can achieve a similar learning accuracy.}
	\label{fig:histogram_AE005}
\end{figure}

{\em Eavesdropping via collective-attack.}---Next, we analyze the CNN learning by assuming an eavesdropping by $\mathscr{L}_E$. To explore a worst-case scenario, we assume a collective attack strategy ${\cal S}_E({\cal P})$ by $\mathscr{L}_E$, and $\eta^{\star} \simeq 0.11$~\cite{Bocquet2011}. Here, we set $\abs{\Theta_Q} = \abs{\Xi_{Q,A}} = \abs{\Xi_{Q,E}}$. Then, the noisy datasets $\Xi_{Q,A}$ and $\Xi_{Q,E}$ for $\mathscr{L}_A$ and $\mathscr{L}_E$ are generated via ${\cal P}$. Both $\mathscr{L}_A$'s and $\mathscr{L}_E$'s CNN learnings are performed on the identical GPU (RTX-3080) workstation. This simulation is repeated $150$ times for each pre-trained CNN model (DN, XC, and NNL) under different $\epsilon_T$ and $\abs{\Theta_Q}$. Here, we consider two cases of $\epsilon_T=0.01$ and $0.03$, and three cases of $\abs{\Theta_Q}=5000$, $10000$, and $15000$. In Fig.~\ref{fig:learning_P}, we plot the learning probabilities with respect to $\eta_A$. When $\eta_A = \eta^{\star} = 0.11$, both $\mathscr{L}_A$ and $\mathscr{L}_E$ have (nearly) same learning probabilities. However, as $\eta_A$ decreases with less aggressive eavesdropping, $\mathscr{L}_A$'s learning probability increases while $\mathscr{L}_E$'s decreases. This pattern indicates a strong correlation between the amount of information extractable from the training data and the learning probability. These observations align well with our {\bf Theorem~\ref{thm:2}} and {\bf Conjecture~\ref{conj:1}}. To more clearly observe the differences in learning outcome quality between $\mathscr{L}_A$ and $\mathscr{L}_E$, we compare the histogram distributions of $\epsilon$ achieved from the learning of each CNN model (DN, XC, and NNL). Herein, we draw the histograms for three different $\eta_A$ values, $0.01$, $0.03$ and $0.05$,  in Fig.~\ref{fig:histogram_AE001}, Fig.~\ref{fig:histogram_AE003}, and Fig.~\ref{fig:histogram_AE005}, respectively. These histograms show that in $\mathscr{L}_A$'s learning, the accuracy below the desired threshold $\epsilon_T$ can consistently be achieved, while the accuracies obtained in $\mathscr{L}_E$'s learning are inconsistent. Such the performance limitations of $\mathscr{L}_E$ are evident when $\mathscr{L}_A$'s dataset $\Xi_{Q,A}$ is less noisy, or equivalently, when $\eta_A$ is small. The results demonstrate that $\mathscr{L}_A$ can achieve a guaranteed superior learning quality. However, with significant performance differences and/or for excessively large data sizes $\abs{\Theta_Q}$, the results may deviate from the predictions. More specifically, this includes cases where the model is robust enough to maintain its good learning performance regardless of the noises; or cases where the dataset is so large that even a high fraction of the contaminated data does not significantly impact the learning. For example, in the performance-optimized model with ample data, the reduction in $\mathscr{L}_E$'s learning probability in the noise levels $\eta_A \in (0.05, \eta^{\star}]$ is not particularly abrupt, as shown by the $\mathscr{L}_E$'s NNL results in Figs.~\ref{fig:learning_P}(c) and~\ref{fig:learning_P}(f), as well as Fig.~\ref{fig:histogram_AE005}. For the DN or XC models with large training data and $\eta_A$ close to $\eta^\star$, the learning probabilities of $\mathscr{L}_E$ appear similar to, or even higher than, those of $\mathscr{L}_A$. This phenomenon reflects a heuristic trait of the machine learning.


\section{Conclusion}

In this study, we have explored the conditions to ensure a specific quality of learning outcomes for an authorized learner, building on quantum label encoding and secure data transformation. Unlike previous studies, we identified the precise conditions under which only an authorized learner can achieve superior learning results. By establishing a connection between the probably-approximately-correct (PAC) learning and the concept of learning probability, we provided a quantitative measure of the learner's performance. Our findings demonstrated that under certain conditions, an authorized learner can attain a guaranteed learning quality, while an eavesdropper is not ensured the same results. However, these conditions do not entirely prevent an eavesdropping learner from obtaining some level of learning quality, which reflects the inherent heuristic nature of machine learning and the tightness of the PAC sample-complexity bound when dealing with noisy data. If a perfectly tight sample-complexity bound can be found---still underived in the computational machine learning---our findings would directly lead to a stringent condition that strictly prevents the eavesdropping learner(s) from surpassing a specific threshold of learning quality.

Beyond theoretical proofs, we applied our approach to practical scenarios, the image classification, using convolutional neural networks (CNNs). By incorporating the quantum label encoding and data transmission protocol into the CNN-based image classification, we validated our main theoretical prediction: a specific level of the learning accuracy ($\epsilon$) and confidence ($1-\delta$) can be guaranteed exclusively for the authorized learner. This empirical evidence highlights the practical relevance of our theoretical results in real-world applications. Here it should be emphasized the authorized learner's superior outcomes and the data security can be ensured purely through the authorized-learner-only measurable quantities related to the training data, namely, its size ($\abs{\Xi_{Q,A}}$) and noise degree ($\eta_A$).

Our study introduces an alternative paradigm that integrates quantum computing, machine learning, and data security. By employing the quantum encoding, our approach ensures an exclusive learning performance for the authorized user, enhancing the data security. These findings have the potential to expand the researches of quantum computing and quantum machine learning, particularly in fields where the data security is crucial.

\section*{Acknowledgements}

J.B. thanks Dr. I. Sohn and Dr. K. Bae for discussions and comments. This work was supported by the Ministry of Trade, Industry, and Energy (MOTIE), Korea, under the project ``Industrial Technology Infrastructure Program'' (RS-2024-00466693), the Institute of Information and Communications Technology Planning and Evaluation (IITP) ``Development of twin field and continuous variable quantum key distribution (QKD) system technology'' (RS-2024-00396999), and the National Research Foundation of Korea (NRF-2023M3K5A1094805, NRF-2023M3K5A1094813, RS-2023-00281456, and RS-2024-00432214). Y.S.Kim also acknowledges the support of the KIST institutional program (2E32941). J.B. thanks to Prof. J. Kim for fruitful discussions. J.B. also appreciate the support from the School of Physics and Quantum Universe Center (QUC), Korea Institute for Advanced Study (KIAS) (Project No.~QP014601). W. S. is supported by the Grant No.~K24L4M1C2 at the Korea Institute of Science and Technology Information (KISTI).

\section*{References}

\bibliographystyle{iop}

\begin{thebibliography}{10}

\bibitem{Biamonte2017}
Biamonte J, Wittek P, Pancotti N, Rebentrost P, Wiebe N and Lloyd S 2017 {\it
  Nature\/} {\bf 549} 195

\bibitem{Ciliberto2018}
Ciliberto C, Herbster M, Ialongo A~D, Pontil M, Rocchetto A, Severini S and
  Wossnig L 2018 {\it Proceedings of the Royal Society A: Mathematical,
  Physical and Engineering Sciences\/} {\bf 474} 20170551

\bibitem{Rebentrost2014}
Rebentrost P, Mohseni M and Lloyd S 2014 {\it Physical review letters\/} {\bf
  113} 130503

\bibitem{Schuld2016}
Schuld M, Sinayskiy I and Petruccione F 2016 {\it Physical Review A\/} {\bf 94}
  022342

\bibitem{Wang2017}
Wang G 2017 {\it Physical review A\/} {\bf 96} 012335

\bibitem{Lloyd2014}
Lloyd S, Mohseni M and Rebentrost P 2014 {\it Nature physics\/} {\bf 10} 631

\bibitem{Aaronson2015}
Aaronson S 2015 {\it Nature Physics\/} {\bf 11} 291

\bibitem{Arunachalam2015}
Arunachalam S, Gheorghiu V, Jochym-O’Connor T, Mosca M and Srinivasan P~V
  2015 {\it New Journal of Physics\/} {\bf 17} 123010

\bibitem{Tang2021}
Tang E 2021 {\it Physical Review Letters\/} {\bf 127} 060503

\bibitem{Schuld2019}
Schuld M and Killoran N 2019 {\it Physical review letters\/} {\bf 122} 040504

\bibitem{Havlivcek2019}
Havl{\'\i}{\v{c}}ek V, C{\'o}rcoles A~D, Temme K, Harrow A~W, Kandala A, Chow
  J~M and Gambetta J~M 2019 {\it Nature\/} {\bf 567} 209

\bibitem{Lloyd2020}
Lloyd S, Schuld M, Ijaz A, Izaac J and Killoran N 2020 {\it arXiv preprint
  arXiv:2001.03622\/}

\bibitem{Weigold2021}
Weigold M, Barzen J, Leymann F and Salm M 2021 {\it IET Quantum
  Communication\/} {\bf 2} 141

\bibitem{Bang2015}
Bang J, Lee S~W and Jeong H 2015 {\it Quantum Information Processing\/} {\bf
  14} 3933

\bibitem{Sheng2017}
Sheng Y~B and Zhou L 2017 {\it Science Bulletin\/} {\bf 62} 1025

\bibitem{Liu2018}
Liu N and Rebentrost P 2018 {\it Physical Review A\/} {\bf 97} 042315

\bibitem{Du2021}
Du Y, Hsieh M~H, Liu T, Tao D and Liu N 2021 {\it Physical Review Research\/}
  {\bf 3} 023153

\bibitem{Llorens2024}
Llorens S, Sent{\'\i}s G and Mu{\~n}oz-Tapia R 2024 {\it Quantum\/} {\bf 8}
  1452

\bibitem{Song2021a}
Song W, Lim Y, Kwon H, Adesso G, Wie{\'s}niak M, Paw{\l}owski M, Kim J and Bang
  J 2021 {\it Physical Review A\/} {\bf 103} 042409

\bibitem{Harney2022}
Harney C and Pirandola S 2022 {\it PRX Quantum\/} {\bf 3} 010311

\bibitem{Valiant1984}
Valiant L~G 1984 {\it Communications of the ACM\/} {\bf 27} 1134

\bibitem{Langley1996}
Langley P 1996 {\it Elements of machine learning\/} (Morgan Kaufmann)

\bibitem{Lee2019}
Lee J~S, Bang J, Hong S, Lee C, Seol K~H, Lee J and Lee K~G 2019 {\it Physical
  Review A\/} {\bf 99} 012313

\bibitem{Song2021b}
Song W, Wie{\'s}niak M, Liu N, Paw{\l}owski M, Lee J, Kim J and Bang J 2021
  {\it Quantum Information Processing\/} {\bf 20} 275

\bibitem{Kotsiantis2011}
Kotsiantis S 2011 {\it Artificial intelligence review\/} {\bf 42} 157

\bibitem{Angluin1994}
Angluin D and Slonim D~K 1994 {\it Machine Learning\/} {\bf 14} 7

\bibitem{Liu2023}
Liu J, Hann C~T and Jiang L 2023 {\it Physical Review A\/} {\bf 108} 032610

\bibitem{Liu2024}
Liu J and Jiang L 2024 {\it IEEE Network\/}

\bibitem{Fuchs1996}
Fuchs C~A and Peres A 1996 {\it Physical Review A\/} {\bf 53} 2038

\bibitem{Fuchs1997}
Fuchs C~A, Gisin N, Griffiths R~B, Niu C~S and Peres A 1997 {\it Physical
  Review A\/} {\bf 56} 1163

\bibitem{Bocquet2011}
Bocquet A, All{\'e}aume R and Leverrier A 2011 {\it Journal of Physics A:
  Mathematical and Theoretical\/} {\bf 45} 025305

\bibitem{Scarani2005}
Scarani V, Iblisdir S, Gisin N and Ac{\'\i}n A 2005 {\it Reviews of Modern
  Physics\/} {\bf 77} 1225

\bibitem{Dang2007}
Dang G~F and Fan H 2007 {\it Physical Review A—Atomic, Molecular, and Optical
  Physics\/} {\bf 76} 022323

\bibitem{Banaszek2001}
Banaszek K 2001 {\it Physical Review Letters\/} {\bf 86} 1366

\bibitem{Devetak2005}
Devetak I 2005 {\it IEEE Transactions on Information Theory\/} {\bf 51} 44

\bibitem{Cai2004}
Cai N, Winter A and Yeung R~W 2004 {\it problems of information transmission\/}
  {\bf 40} 318

\bibitem{Holevo2011book}
Holevo A~S 2011 {\it Probabilistic and statistical aspects of quantum theory\/}
  vol.~1 (Springer Science \& Business Media)

\bibitem{Schumacher1997}
Schumacher B and Westmoreland M~D 1997 {\it Physical Review A\/} {\bf 56} 131

\bibitem{Chen2016}
Chen X, Zhang Y et~al. 2016 {\it Journal of Machine Learning Research\/} {\bf
  17} 1

\end{thebibliography}

\end{document}